\pdfoutput=1
\documentclass[letterpaper, 10 pt, conference]{ieeeconf}  

\IEEEoverridecommandlockouts                              

\title{\LARGE \bf
  Collision-aware Task Assignment for Multi-Robot Systems
}

\author{Fang Wu$^{1}$, Vivek Shankar Varadharajan$^{1}$ and Giovanni Beltrame$^{1}$
\thanks{*This work was not supported by any organization}
\thanks{$^{1}$Fang Wu, Vivek Shankar Varadharajan and Giovanni Beltrame are with the Department of
  Computer and Software Engineering, Polytechnique Montreal, Canada.
        {\tt\small giovanni.beltrame@polymtl.ca}}%
}

\usepackage{graphicx}
\usepackage[linesnumbered,ruled,vlined]{algorithm2e}
\usepackage{caption}
\usepackage{subcaption}
\usepackage{todonotes}

\makeatletter
\newcommand*{\rom}[1]{\expandafter\@slowromancap\romannumeral #1@}
\makeatother

\usepackage[belowskip=0pt,aboveskip=2pt]{caption}
\begin{document}

\maketitle
\thispagestyle{empty}
\pagestyle{empty}

\begin{abstract}

  We propose a novel formulation of the collision-aware task assignment (CATA) problem and a decentralized auction-based algorithm to solve the problem with optimality bound. Using a collision cone, we predict potential
  collisions and introduce a binary decision variable into the local reward
  function for task bidding. We further improve CATA by implementing a
  receding collision horizon to address the stopping robot scenario, i.e. when robots are confined to their task location and become static obstacles to
  other moving robots. The auction-based algorithm encourages the robots to
  bid for tasks with collision mitigation considerations. We validate the
  improved task assignment solution with both simulation and experimental
  results, which show significant reduction of overlapping paths as well as deadlocks.

\end{abstract}

\section{INTRODUCTION}

A successful multi-robot mission generally requires the robots in a team to
collaborate towards a common goal, with each robot performing task in
coordination with its teammates. Distributing tasks among robots is commonly
referred as \emph{task assignment problem}, which is generally followed by
solving a multi-robot path planning problem to route every robot to its
assigned task. In any realistic scenario, path planning should consider
collision avoidance~\cite{Breitenmoser2016,Leonard2017,Berg2011}.

Task assignment is a well-known combinatorial optimization problem: $N_T$
tasks need to be assigned to $N_R$ robots to minimize a global cost
function. This global cost function depends on the specific mission and
customized constraints are usually considered for the optimization
problem. The task assignment problem is NP-hard, requiring heuristic
approaches \cite{PILLAC20131,ritzinger:hal-01224562}. Search in the solution
space is usually performed in a centralized setting, where the robots need to
communicate with a central planner. This reduces the computation complexity on
individual robots, but introduces a single point of failure in the
mission. Data transmission towards the central planner also increases with the
number of robots, causing congestion. Some decentralized
methods~\cite{Shima2007} instantiate a task planner on each robot and make use
of consensus algorithms to reach a consistent representation of the
environment before task assignment. Other methods~\cite{Choi2009} let
individual robots conduct local planning first and then exploit the consensus
algorithm to achieve agreement on the assignment.

Task assignment procedures generate a collection of tuples $(r_i, t_l)$, where
$r_i\in{1,...,N_R}$ refers to the robot ID and $t_l\in{1,...,N_T}$ refers to a
task identifier. The subsequent multi-robot path planning problem aims at finding
$\textbf{p}_{il} (r_i, t_l)$, the optimal path from the location of robot
$r_i$ to the location of task $t_l$. Global planners~\cite{Honig2018} use
prior information about the environment and predict potential collisions
between robots, and typically assume perfect motion execution and minimal
external disturbance. On the contrary, reactive collision avoidance
strategies~\cite{Berg2011} naturally work in a decentralized system and are
more robust to a dynamic environment. However, all these solutions deal with
predefined task assignment, despite the clear indication that task assignment
and path planning are inherently coupled. Recent work~\cite{Alonso-Mora2018}
has confirmed that higher level reasoning is necessary to improve the
collision avoidance performance as $N_R$ increases. Integrating collision
awareness with the task assignment process could potentially boost the
performance of path planning as well as collision avoidance algorithms.

Existing studies that integrate task assignment and path planning are very
limited. Cons and Edison et al.~\cite{Cons2014,Edison2011} explored the coupled nature of
task assignment and path optimization by considering the actuation constraints
of fixed-wing UAVs, but assumed altitude layering for multi-robot path
planning, which means each aircraft is flying at a different altitude. This
would soon become unrealistic as $N_R$ increases. Yao et al.~\cite{Yao2019}
implemented a reestimation mechanism to reject the task assignment results
when unrealistic task completion time occurs during the path planning
stage. To the best of our knowledge, no existing work has considered collision
mitigation within the task assignment problem. This is due to the common
assumption that tasks are distributed sparsely and robot dimensions are
significantly small relative to path lengths. This assumption becomes invalid
for operations in task-dense environments, such as those involving automatic
construction or collective transportation.  When considering a congested
scenario where multiple robots attempt to cross paths with each other to reach
their assignments, neglecting collision mitigation during the task assignment
stage causes the mission completion time/ travel distance to increase without
upper bound, or even worse, robots to collide with each other as the scenario
becomes too complex for local collision avoidance strategies.

Our work formulates a task assignment problem with collision mitigation terms
and develops a decentralized task assignment method that makes use of
consensus algorithms. In this paper, we present an analytical derivation that
provides a guaranteed optimality lower bound, and validate our results with
simulation and experimental campaigns.

The remainder of this paper is organized as follows: Section~\ref{section_two}
presents the classical task assignment problem, collision cones, and Buzz
\cite{Pinciroli2015} (a language designed for robot swarms) as preliminaries
to the following discussion; Section~\ref{section_three} provides the
mathematical formulation of the problem and proposes a solution to the
collision-aware task assignment problem; Sections~\ref{section_four} and
\ref{section_five} present simulation and experimental results on the
performance of our method; finally, Section~\ref{section_six} offers some
concluding remarks.

\section{PRELIMINARIES}
\label{section_two}

\subsection{Classical task assignment problem}

The objective of a task assignment problem is to maximize a global scoring function or to minimize a global cost function while enforcing a set of
constraints. In this work, we consider the maximization problem and express the global scoring function as a sum of local reward functions:
\begin{equation}
\label{equ_1}
    max \sum_{r_i=1}^{N_R}\sum_{t_l=1}^{N_T}b_{il}x_{il}
\end{equation}
subject to 
$$
    \sum_{r_i=1}^{N_R} x_{il} \leq 1 \hspace{1cm} \forall t_l \in {1,...,N_T}
$$
$$
    \sum_{t_l=1}^{N_T} x_{il} \leq L_T \hspace{1cm} \forall r_i \in {1,...,N_R}
$$
$$
    x_{il} \in \{0, 1\} \hspace{0.5cm}\forall r_i \in {1,...,N_R} \hspace{0.5cm}\forall t_l \in {1,...,N_T}
$$
where $b_{il}$ is the local reward that occurs when assigning robot $r_i$ to
task $t_l$; $x_{il} \in \{0,1\}$ indicates whether robot $r_i$ is assigned to,
or in market-based strategy parlance, has won the task $t_l$; $N_T$ refers to
the total number of tasks, and $N_R$ refers to the total number of
robots. $\sum_{r_i=1}^{N_R} x_{il} \leq 1, \forall t_l \in {1,...,N_T}$
indicates that no two robots are assigned to the same task. $L_T$ represents
the maximum number of assignments for each robot.  A special case of the
formulation above is when $L_T = 1$, which is commonly referred to as the
single-assignment problem.

Choi et al.~\cite{Choi2009} proposed a consensus-based decentralized auction
algorithm for both the single and multi-assignment problems. Under the
\textit{diminishing marginal gain} assumption on the global scoring function, the
algorithm guarantees convergence and solution optimality with a lower
bound. The \textit{diminishing marginal gain} condition assumes the local
reward of a task does not increase as other tasks are being assigned before
it, which is true for many reward functions used in search and exploration
problems.  However, in~\cite{Choi2009} and its followup
work~\cite{Buckman2018}, the local reward of robot $r_i$ winning task $t_l$
only depends on its own previously-won tasks in a multi-assignment
problem. For the single-assignment problem, the local reward function is
static. In this paper, we formulate the single-assignment problem with
collision mitigation, meaning the local reward of robot $r_i$ winning
any task also depends on which task its neighbor $r_j$ has won.

\subsection{Collision cone}

\begin{figure}
      \centering
      
      \includegraphics[width=.6\columnwidth]{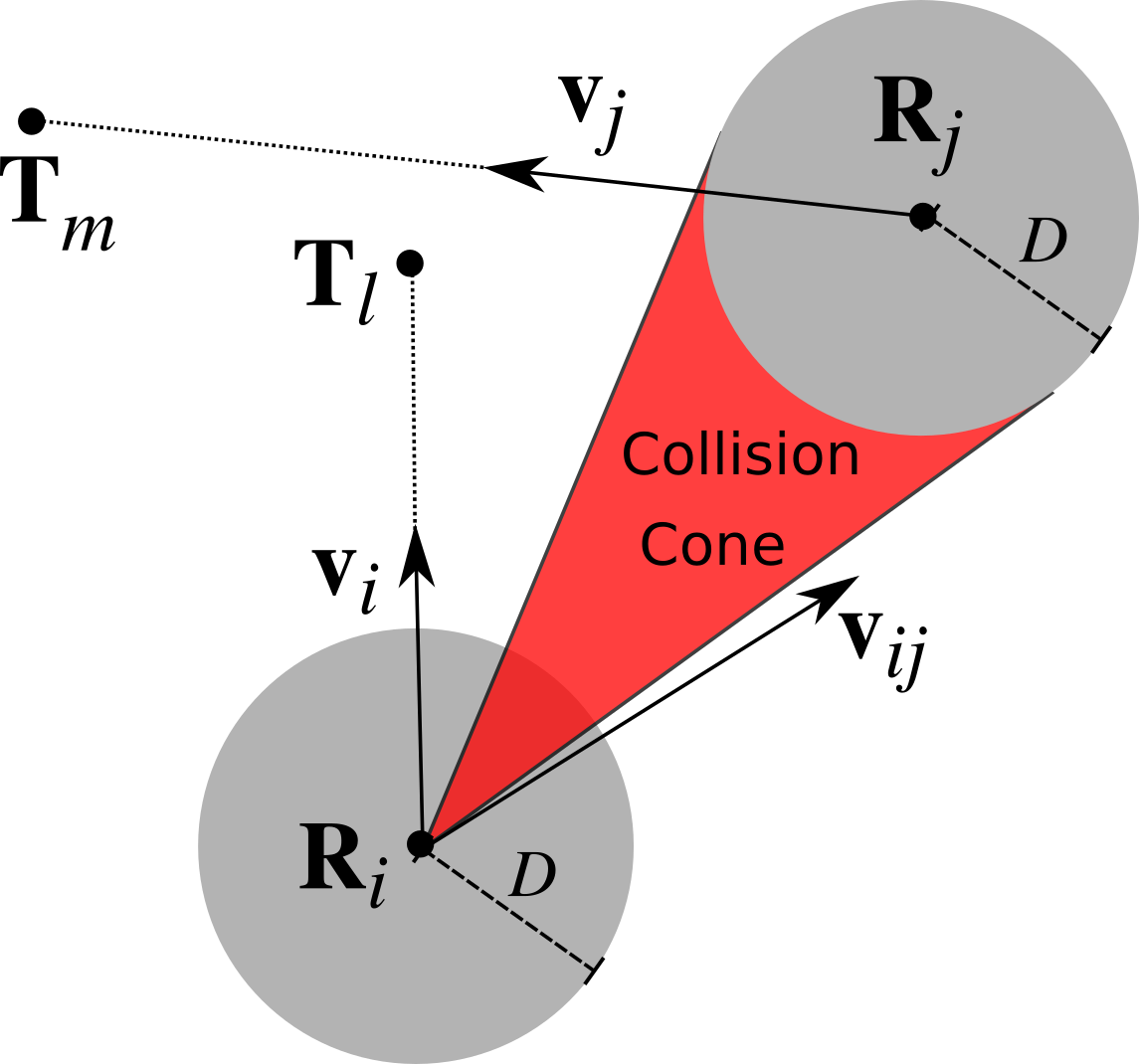}
      \caption{Collision cone regarding two moving robots}
      \label{fig_collisioncone}
   \end{figure}

The collision cone has been widely used to predict collisions between two moving
robots from based on their current locations and velocities. As depicted in
Fig.~\ref{fig_collisioncone}, when robot $r_i$ at location $\textbf{R}_i$ is moving with velocity
$\textbf{v}_i$ and robot $r_j$ at location $\textbf{R}_j$ is moving with
velocity $\textbf{v}_j$, a corresponding collision cone $C_{ij}$ can be
generated with a predefined safety distance $D$, which is indicated by the
grey circular area. By determining whether the relative velocity
$\textbf{v}_{ij}$ lies inside or outside the collision cone, we can predict
potential future collisions. Task locations $\textbf{t}_l$ and $\textbf{t}_m$
are also indicated in Fig.1 with the underlying assumption that the robots are
capable of moving towards the task locations in a straight line regardless of
their orientations. Extensive studies~\cite{Wilkie2009, Snape2010, Lalish2009}
have been integrating the actuation constraints into the collision avoidance
methods based on collision cones. In this paper, the robots are assumed to be
holonomic for the sake of simplicity although the formulation proposed in Section
\ref{section_three} does not rely on this assumption.
   


\section{COLLISION-AWARE TASK ASSIGNMENT}
\label{section_three}

\subsection{Problem statement}
\label{section_three_sub_a}

Given a set of $N_T$ tasks and a set of $N_R$ robots, the goal is to find an
optimal assignment that maximizes a global scoring function while minimizing
potential collision incidents. Each robot can be assigned to one task at most
and no two robots should be assigned to the same task. The mathematical form
of the problem can be written as below:
\begin{equation}
    max \sum_{r_i=1}^{N_R}\sum_{t_l=1}^{N_T}b_{il}x_{il} - \sum_{r_i=1}^{N_R}\sum_{t_l=1}^{N_T}\sum_{r_j=1}^{N_R}\sum_{t_m=1}^{N_T}c_{ijlm}x_{il}x_{jm}
\label{equ_2}
\end{equation}
subject to 
$$
    \sum_{t_l=1}^{N_T} x_{il} \leq 1 \hspace{0.5cm} \sum_{t_m=1}^{N_T} x_{jm} \leq 1 \hspace{0.5cm} \forall r_{i,j} \in {1,...,N_R}
$$
$$
    \sum_{r_i=1}^{N_R} x_{il} \leq 1 \hspace{0.5cm} \sum_{r_j=1}^{N_R} x_{jm} \leq 1 \hspace{0.5cm} \forall t_{l,m} \in {1,...,N_T}
$$
$$
    x_{il}, x_{jm} \in \{0, 1\} \hspace{0.5cm} \forall r_{i,j} \in {1,...,N_R} \hspace{0.5cm}\forall t_{l,m} \in {1,...,N_T}
$$
$$
    i \neq j, \hspace{1.0cm} l \neq m
$$
where $x_{jm}$ indicates the assignment status of robot $r_j$, a neighbor of
robot $r_i$. And $c_{ijlm}$ is the cost of assigning robot $r_i$ to task $t_l$
when robot $r_j$ is already assigned to task $t_m$. $b_{il}$, similar to the
term in (1), is the local reward that occurs when assigning robot $r_i$ to
task $t_l$, which is independent of the assignment of robot $r_j$. It is worth
noting (\ref{equ_2}) resembles the general form of the quadratic assignment
problem~\cite{Burkard}, where the inequality constraints become
equality constraints. The quadratic assignment problem is one of the fundamental
combinatorial optimization problems (from the facilities
location family) and to the best of our knowledge this analogy has not been
drawn before.

By rearranging the terms in (\ref{equ_2}), the optimization function can be
written as
\begin{equation}
    max \sum_{r_i=1}^{N_R}\sum_{t_l=1}^{N_T}b_{il}x_{il}( 1- \sum_{r_j=1}^{N_R}\sum_{t_m=1}^{N_T}\frac{c_{ijlm}}{b_{il}} x_{jm}).
\end{equation}
To simplify the notation, we replace the term $\frac{c_{ijlm}}{b_{il}}$ with a
collision weight $w_{ijlm}$ that is determined by the collision status when
robot $r_i$ is assigned to task $t_l$ and robot $r_j$ to task $t_m$.
\begin{equation}
\label{equ_4}
    max \sum_{r_i=1}^{N_R}\sum_{t_l=1}^{N_T}b_{ijlm}x_{il}
\end{equation}
where 
\begin{equation}
\label{equ_5}
    b_{ijlm} = (1- \sum_{r_j=1}^{N_R}\sum_{t_m=1}^{N_T}w_{ijlm} x_{jm})b_{il}
\end{equation}
The collision weights can assume different values for each robot-task pair and
it can be deduced that $\sum_{t_m=1}^{N_T}w_{ijlm} \in [0,1] $,
$\forall r_i \in 1,...,N_T$ as negative scores do not have real-world meaning:
the worst case scenario is that path overlapping prevents the robot from
reaching its assigned task, leading to a null reward. Therefore, we can
approximate the sum in (\ref{equ_5}) with a binary decision variable
$W_{ijlm}$,
\begin{equation}
\label{equ_6}
    b_{ijlm} = (1- W_{il})b_{il}
\end{equation}
This binary variable is determined by the collision status when robot $r_i$ is
assigned to task $t_l$:
\begin{equation}
    W_{il} = \left\{ \begin{array}{l}
                0 \hspace{0.5cm} if \hspace{0.2cm} \textbf{v}_{ij} \notin C_{ij} \hspace{0.5cm} \forall r_j \in 1,...,N_R \\
                1 \hspace{0.5cm} otherwise\\
                \end{array} \right.
\end{equation}
where $\textbf{v}_{ij}$ refers to the relative velocity as depicted in Fig.~\ref{fig_collisioncone},  $C_{ij}(\textbf{R}_i, \textbf{R}_j, D)$ refers to the collision cone determined by the robots' location and predefined safety distance. 

By comparing (\ref{equ_4}) with the classical task assignment formulation
(\ref{equ_1}), it is apparent now that the optimization function has similar
form while the reward function is multiplied by a binary decision variable. As
more tasks are assigned before robot $r_i$ bids for task $t_l$, the binary
decision variable can only increases from zero to one, which would only
further reduce the local reward. We can then draw a conclusion: when
local reward function $b_{il}$ satisfies the \textit{diminishing marginal
  gain} condition, the reward function $b_{ijlm}$ that considers collision
mitigation also satisfies the \textit{diminishing marginal gain}
condition. This allows us to take advantages of properties that are already
proven in \cite{Choi2009}, which guarantees 50\% optimality assuming all the
robots have accurate knowledge of the situation. We give detailed proof of optimality lower bound in the Appendix.

\subsection{Auction and consensus strategy implemented in Buzz}
Here we implement the auction and consensus strategy in
Buzz~\cite{Pinciroli2015} and use the virtual stigmergy structure as the
information propagation infrastructure for consensus agreements. Virtual
stigmergy~\cite{Pinciroli2016b} is a (key,value) pair based shared memory that
allows the robots to globally agree on the values of a set of variables, which
is a good fit for the operation here.

The consensus based auction algorithm consists of two phases: the auction process
and the consensus process.  During the auction process, each robot first fetches the
assignment set $A$ from the virtual stigmergy. The assignment set consists of
tuples that link the robot ID to its assigned task identifier,
$A_j: (r_j, t_m)$. This set can be searched with either the robot ID or the
task identifier, $A(r_j) \to t_m$, $A(t_m) \to r_j$. Then, the robot
computes its own bid for every task that has not been assigned, while
considering the respective collision status with each of its neighbors that
have already won a task. Finally, each robot determines its own highest bid as a tuple
$B_i: (r_i, b_{max})$ and puts this tuple in the virtual stigmergy, which
is akin to broadcasting this tuple to its
neighbors. Algorithm~\ref{algo_1} shows the procedure of robot $r_i$'s auction
phase. The time-discounted reward function $S$ is used to compute the local
reward before any collision mitigation consideration:
\begin{equation}
    S(r_i, t_l) = \lambda_l^{\tau_i^l}V_l
\end{equation}
where $\lambda_l \leq 1$ is the discounting factor for task $t_l$ and
$\tau_i^l$ is the estimated travel time for robot $r_i$ to reach task
$t_l$. $V_l$ refers to the inherent value of task $t_l$, which usually depends
on the importance of specific task to the whole mission. For the sake of
simplicity,  this paper considers a similar discounting factor and inherent value for all the tasks.

In the event of simultaneous modification of the global bidding tuple in the virtual stigmergy, we resolve the conflict by accepting the tuple with higher bid. After every robot has submitted its bid tuple, each robot updates the assignment set $A$ accordingly. Algorithm~\ref{algo_2} details the update policies on
robot $r_i$.

\begin{algorithm}
\DontPrintSemicolon 
\KwIn{Assignment set $A=\{A_1, A_2, \ldots, A_{N_R}\}$}
\KwOut{Bidding tuple}
\For{$p \gets 1$ \textbf{to} $N_T$} {
    $b_{ip} = S(r_i, t_p)$
    
    \For{$q \gets 1$ \textbf{to} $N_R$} {
      \If{$W_{ip} = 1$}{
        \Return{}
        }
      \If{$q \neq r_i$} {
        \If{$A_q \in A$}{
            $\textbf{v}_i = norm(\textbf{T}_p-\textbf{R}_i)$
            $\textbf{v}_q = norm(\textbf{T}_{A(q)}-\textbf{R}_q)$
            
            \If {$\textbf{v}_i-\textbf{v}_q \in C(\textbf{R}_i, \textbf{R}_q, D)$}{
                $W_{ip} = 1$
            }
            
        }
        
      }
    }
    $b_{ip} = (1-W_{ip})b_{ip}$
}
\Return{$B_i: (r_i, max(b_{ip}))$}\;
\caption{finds the local highest bid on robot $r_i$}
\label{algo_1}
\end{algorithm}

\begin{algorithm}[htpb]
\DontPrintSemicolon 
\KwIn{Global bid tuple $B_g : (r_g, b_g)$}
\KwOut{Updated assignment set $A=\{A_1, A_2, \ldots, A_{N_R}\}$}

\If{$r_g == r_i$}{
$A(r_i) \gets argmax(b_{ip})$
}

\caption{updates the assignment set in virtual stigmergy}
\label{algo_2}
\end{algorithm}

\subsection{Receding collision horizon }
\label{section_three_sub_c}

\begin{figure}

\begin{subfigure}{.25\textwidth}
\captionsetup{width=0.8\textwidth}

 \centering
  \includegraphics[width=0.9\textwidth]{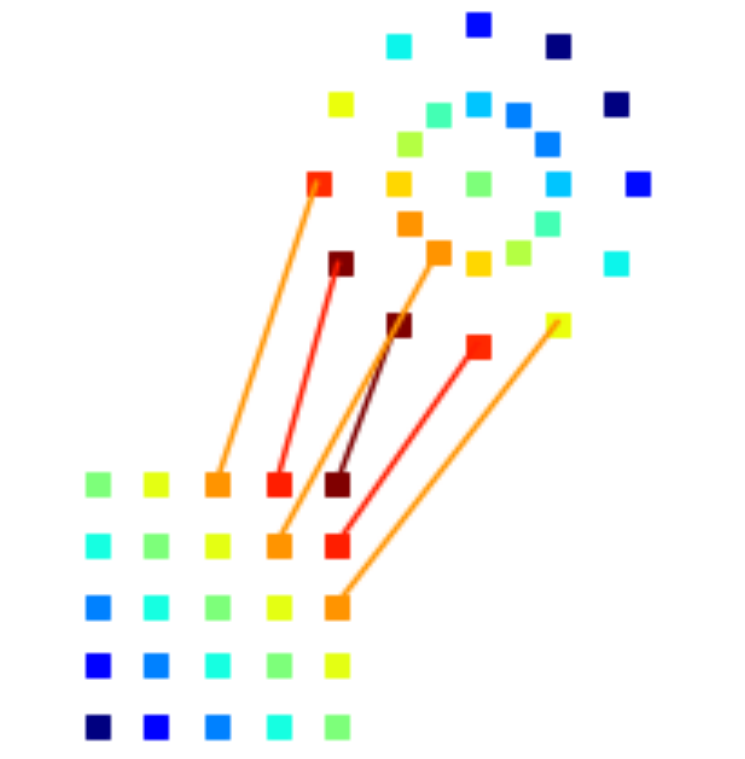}
  \caption{bidding results without collision mitigation}
  \label{fig_receding_horizon_sub1}
\end{subfigure}%
\begin{subfigure}{.25\textwidth}
\captionsetup{width=0.8\textwidth}

\centering
  \includegraphics[width=.9\textwidth]{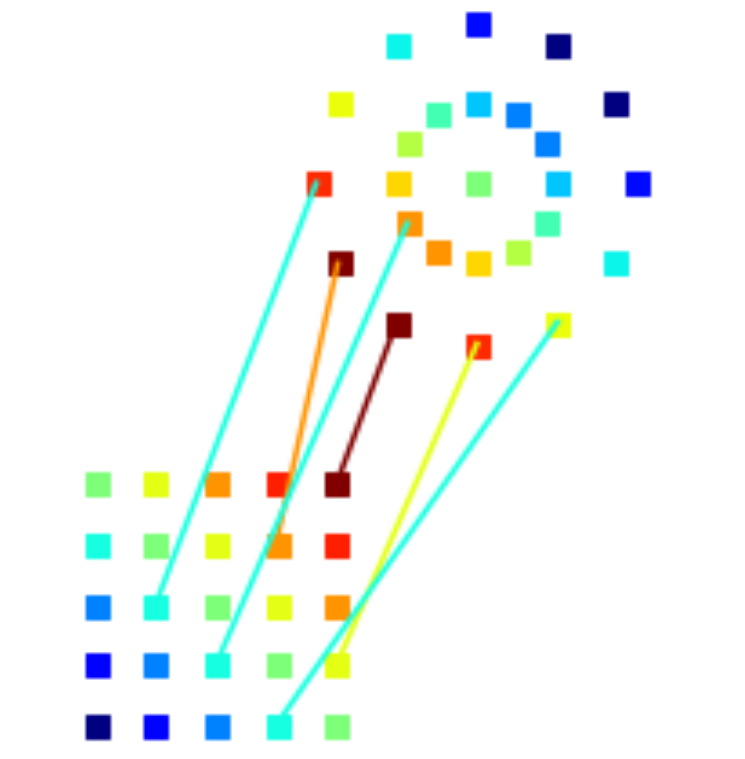}
  \caption{bidding results with collision mitigation}
  \label{fig_receding_horizon_sub2}
\end{subfigure}
\caption{Effect of the receding collision horizon}
\label{fig_receding_horizon}
\end{figure}

By introducing a binary decision variable into the local reward function,
robot $r_i$ is discouraged to bid for task $t_l$ when robot $r_j$ has already
been assigned to task $t_m$ and potential collision is predicted. This would
effectively reduce the crossing path incidents during mission
execution. However, there exists another collision scenario that is often
neglected in existing works: the stopping robot scenario. To better
explain this scenario, we show a task assignment problem with 25
robots and 25 tasks illustrated in Fig.~\ref{fig_receding_horizon}.

The robots are uniformly distributed within a square-shaped arena, referred as
\textit{robot arena}, and the tasks are arranged in three
layers within a circular area, referred as \textit{task area}. The quantity of
robots/tasks as well as their formations are simply chosen for illustration
purpose and do not impose any assumption for following discussion. Jet
colormap is used for coloring. The color of the robots depends on the distance
between specific robot and the center of \textit{task area}, while color of
the tasks depends on the distance between the specific task and the center of
\textit{robot arena}. This paper refers to the robots that are closer to
\textit{task area} as \textit{front-row robots} and robots that are farther from
\textit{task area} as \textit{back-row robots}. The same
terms are also used to describe tasks regarding their distance to the center
of the \textit{robot arena}.  When using a time-discounted reward function during
the auction process without a collision mitigation term, the \textit{front-row
  robots} naturally win the \textit{front-row tasks} and the \textit{back-row
  robots} end up winning the \textit{back-row tasks}, as shown in
Fig.~\ref{fig_receding_horizon_sub1}. The bidding results after six iterations
are indicated with lines connecting the robot with its assigned task.

Although the actual content of the task is outside the scope of this paper, it
is reasonable to assume the robots will be bound to the task location for a
certain duration of time, which means the \textit{front-row robots} become
static obstacles after arriving at the \textit{front-row tasks}. This
significantly complicate the collision scenarios \textit{back-row robots} have
to face when they reach \textit{task area} and unfortunately, introducing a
binary decision variable into the local reward function cannot effectively
mitigate this issue due to the locality of the collision cone. Therefore,
we propose a receding collision horizon: instead of a static safety distance
$D$, we use a diminishing safety distance during the bidding process.

\begin{figure}

\begin{subfigure}{.25\textwidth}
\captionsetup{width=0.8\textwidth}
 \centering
  \includegraphics[width=\textwidth]{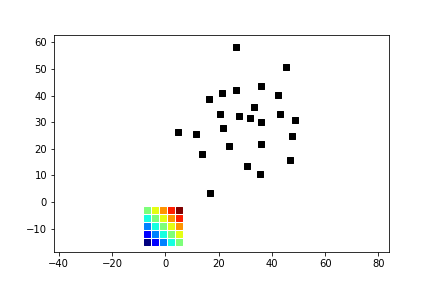}
  \caption{robots are distributed on a grid}
  \label{fig_setup_sub1}
\end{subfigure}%
\begin{subfigure}{.25\textwidth}
\captionsetup{width=0.8\textwidth}
 \centering
  \includegraphics[width=\textwidth]{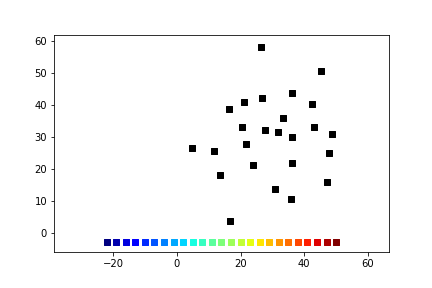}
  \caption{robots are distributed on a line}
  \label{fig_setup_sub2}
\end{subfigure}%
\label{fig_setup}
\caption{Different initial setups}
\end{figure}

When the collision cone is used in collision avoidance algorithms, the safety
distance is usually determined by the robot size as well as its locomotion
capability. This parameter is represented by $D_{min}$ here. In the collision
mitigation context, the interpretation of safety distance $D$ can be further
extended. Intuitively, it is a representation of how far the robot will risk
going into the collision horizon to win a task. Considering a specific
\textit{front-row robot} $r_i$ and its closest neighbor $r_j$, which has been
assigned to task $t_m$: when $D$ is increased so that
$D \geq |\textbf{R}_i-\textbf{R}_j|$, a collision will be predicted between
robot $r_i$ and $r_j$ regardless of robot $r_j$'s assignment. Therefore robot
$r_i$ is discouraged to bid for any task. If we start with a large safety
distance and reduce it only when a zero bid is submitted by all robots,
the \textit{back-row robots} are encouraged to bid for \textit{front-row
  tasks} even when the \textit{front-row robots} are not fully
assigned. Fig.~\ref{fig_receding_horizon_sub2} shows the results after
six bidding iterations when using this scheme.

From the perspective of optimization formulation, a large problem is segmented
into smaller problems after introducing the receding collision horizon. And
each smaller problem still uses the local reward function in the form of
(\ref{equ_6}), which satisfies the \textit{diminishing marginal gain}
condition, thus the 50\% optimality guarantee still holds.

\begin{figure}[h]
	\begin{subfigure}{.25\textwidth}
		\centering
		\includegraphics[width=\textwidth]{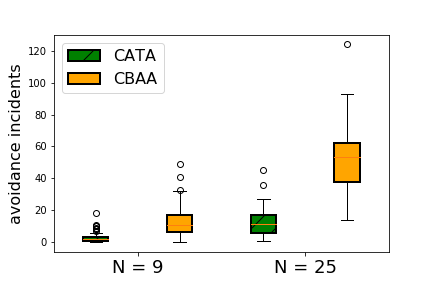}
	\end{subfigure}%
	\begin{subfigure}{.25\textwidth}
		\centering
		\includegraphics[width=\textwidth]{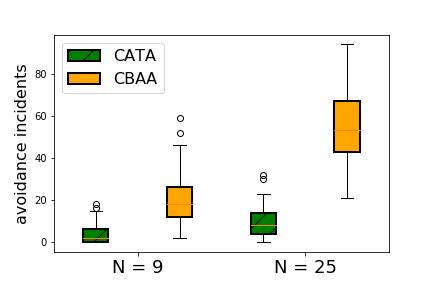}
	\end{subfigure}

	\begin{subfigure}{.25\textwidth}
		\centering
		\includegraphics[width=\textwidth]{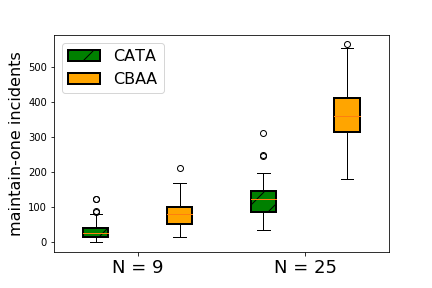}
	\end{subfigure}%
	\begin{subfigure}{.25\textwidth}
		\centering
		\includegraphics[width=\textwidth]{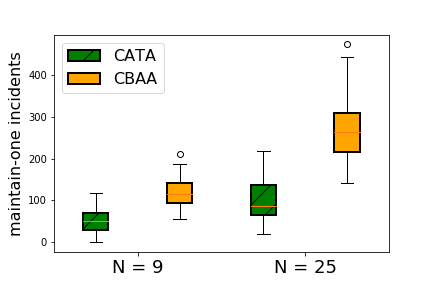}
	\end{subfigure}
	
	\begin{subfigure}{.25\textwidth}
		\centering
		\includegraphics[width=\textwidth]{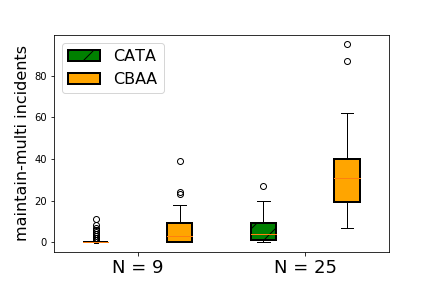}
	
	\end{subfigure}%
	\begin{subfigure}{.25\textwidth}
		\centering
		\includegraphics[width=\textwidth]{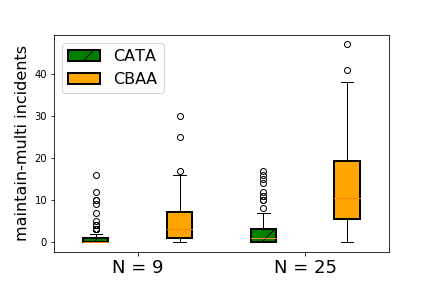}
	\end{subfigure}
	\caption{Box plot of collision incidents: plots of first row shows the \textit{avoidance} incident, second row shows the \textit{maintain-one} incident, third row shows the \textit{maintain-multi} incident; plots of left column refers to the grid setup, right column refers to the line setup }
\label{fig_collision_incidents}
\end{figure}

\section{SIMULATION RESULTS}
\label{section_four}

We developed a greedy algorithm to sequentially find the global bid tuple
$B_g : (r_g, b_g)$ that renders the highest bid given prior assignments. This
centralized procedure generates the same task assignment solution with CATA,
assuming a fully connected network so that every robot is able to submit their
bid before conducting any task selection. Another distributed auction
algorithm for single-assignment with only time-discounted rewards, named
consensus-based auction algorithm (CBAA) in~\cite{Choi2009}, is also simulated
as a control group. We chose the distributed reactive collision avoidance
(DRCA)~\cite{Leonard2017} method as the local collision avoidance strategy due
to its capability of handling multi-robot interaction and maintaining
collision-free before robots even enter the collision cone.  Deconfliction
maintenance is triggered when one robot detects one or multiple neighbors
entering its safety zone, but outside the collision cone. And a deconfliction
maneuver is triggered when one robot detects its neighbor entering its
collision cone.

\begin{figure}	
	\begin{subfigure}{.25\textwidth}
		\centering
		\includegraphics[width=\textwidth]{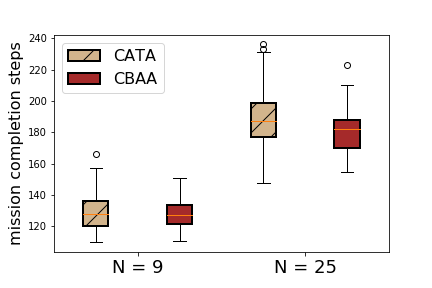}
		\caption{Grid setup}
	\end{subfigure}%
	\begin{subfigure}{.25\textwidth}
		\captionsetup{width=0.8\textwidth}
		\centering
		\includegraphics[width=\textwidth]{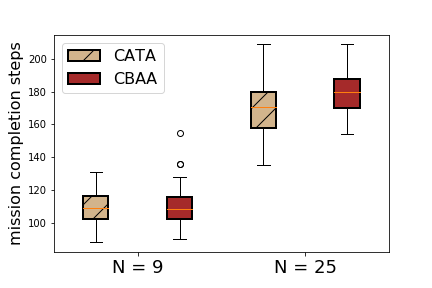}
		\caption{Line setup}
	\end{subfigure}
\caption{Box plot of mission completion steps}
\label{fig_steps}
\end{figure}

\begin{table}[h]
	\caption{Summary of simulation results}
	\label{table_1}
	
	\begin{center}
		\begin{tabular}{ |p{0.15\textwidth}|p{0.05\textwidth}|p{0.05\textwidth}|p{0.05\textwidth}|p{0.05\textwidth}| }
			\hline
			\multicolumn{1}{|c|}{} & \multicolumn{2}{|c|}{Grid setup} & \multicolumn{2}{|c|}{Line setup}\\
			\hline
			& CBAA & CATA & CBAA & CATA\\
			\hline
			Dead-lock for $N=9$   & 9    &2 & 16    &4\\
			
			\hline
			Dead lock for $N=25$  & 52    &8 & 24    &3\\
			\hline
		\end{tabular}
	\end{center}
\end{table}

Here we assume $N_R = N_T$ and use $N$ to replace the notations for the
sake of simplicity, although this is not required for the proposed
algorithm. Regarding the robots' initial locations, we exploited two different
patterns for comparison. Taking $N = 25$ as an example, twenty-five robots are
uniformly located on a 5x5 grid in Fig.~\ref{fig_setup_sub1}, represented by
the colored squares and referred as grid setup in the later part; robots are
uniformly located on a horizontal line in Fig.~\ref{fig_setup_sub2}, referred
as line setup in later part. These two patterns are chosen instead of randomly
generated robot locations, because real-life task assignment scenarios often
deal with initial configurations like these. Also, this comparison can
provide more insights about the relation between collision mitigation
performance and the initial setup. Task locations $\textbf{T}$, are randomly
generated from two normal distributions, $T_x\sim \mathcal{N}(\sigma = 10)$,
$T_y\sim \mathcal{N}(\sigma = 10)$ (black squares).

\begin{figure*}	
    \centering
    \begin{subfigure}{.32\textwidth}
		\captionsetup{width=0.9\textwidth}
		\centering
		\includegraphics[width=\textwidth]{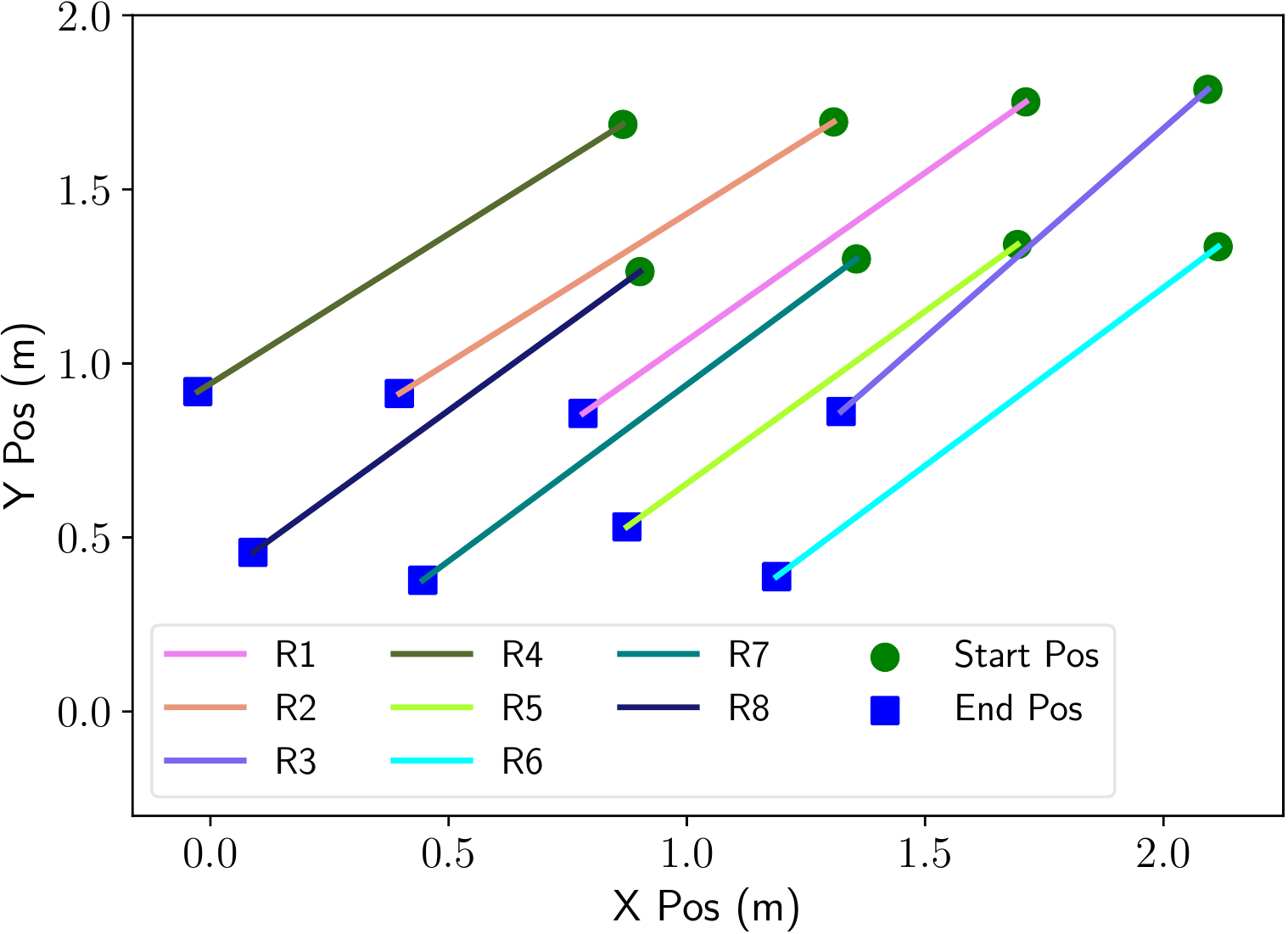}
		\caption{Optimal}
	\end{subfigure}%
	\begin{subfigure}{.32\textwidth}
		\captionsetup{width=0.9\textwidth}
		\centering
        \includegraphics[width=\textwidth]{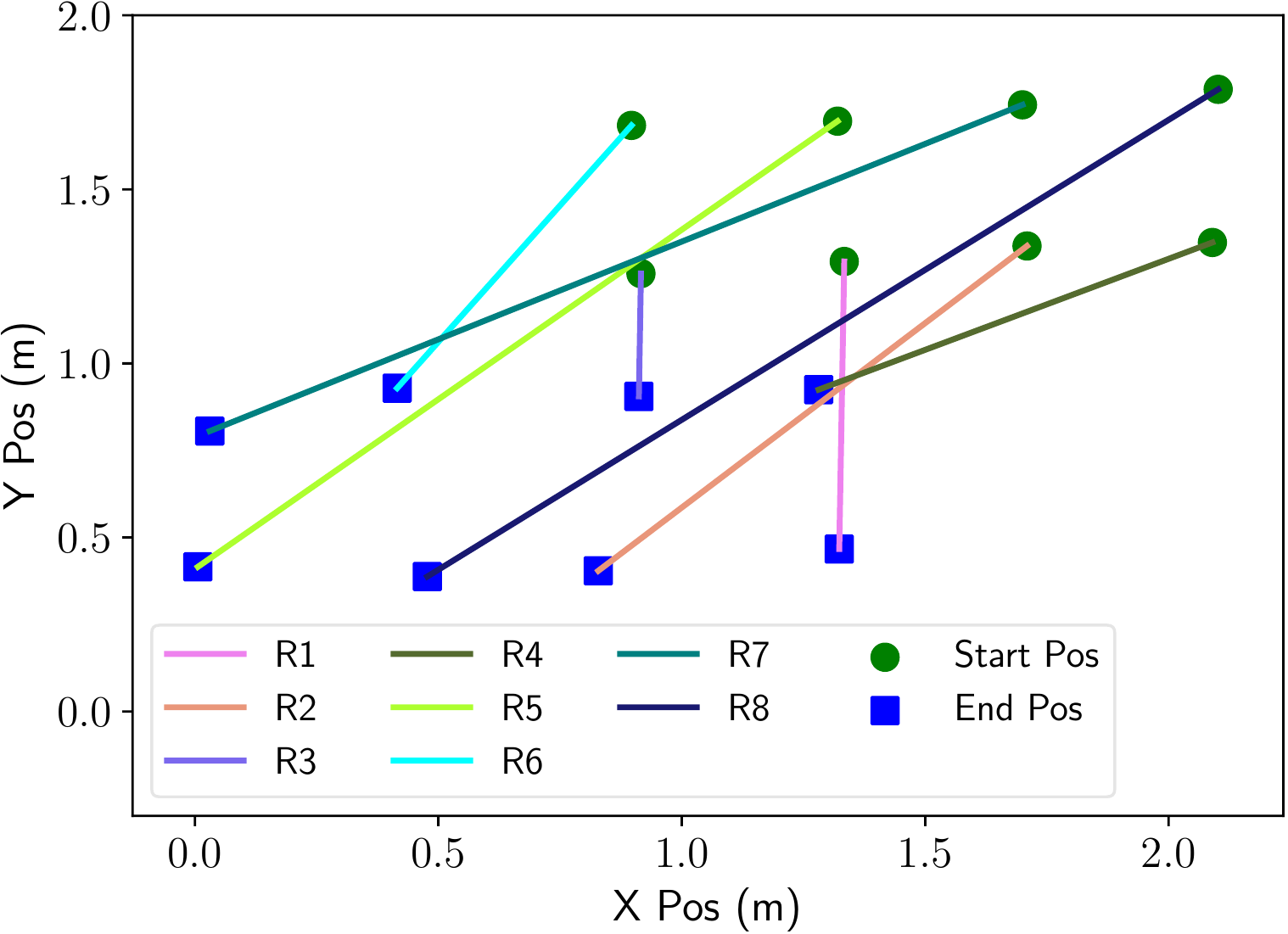}
		\caption{CATA}
	\end{subfigure}
	\begin{subfigure}{.32\textwidth}
		\captionsetup{width=0.9\textwidth}
		\centering
		\includegraphics[width=\textwidth]{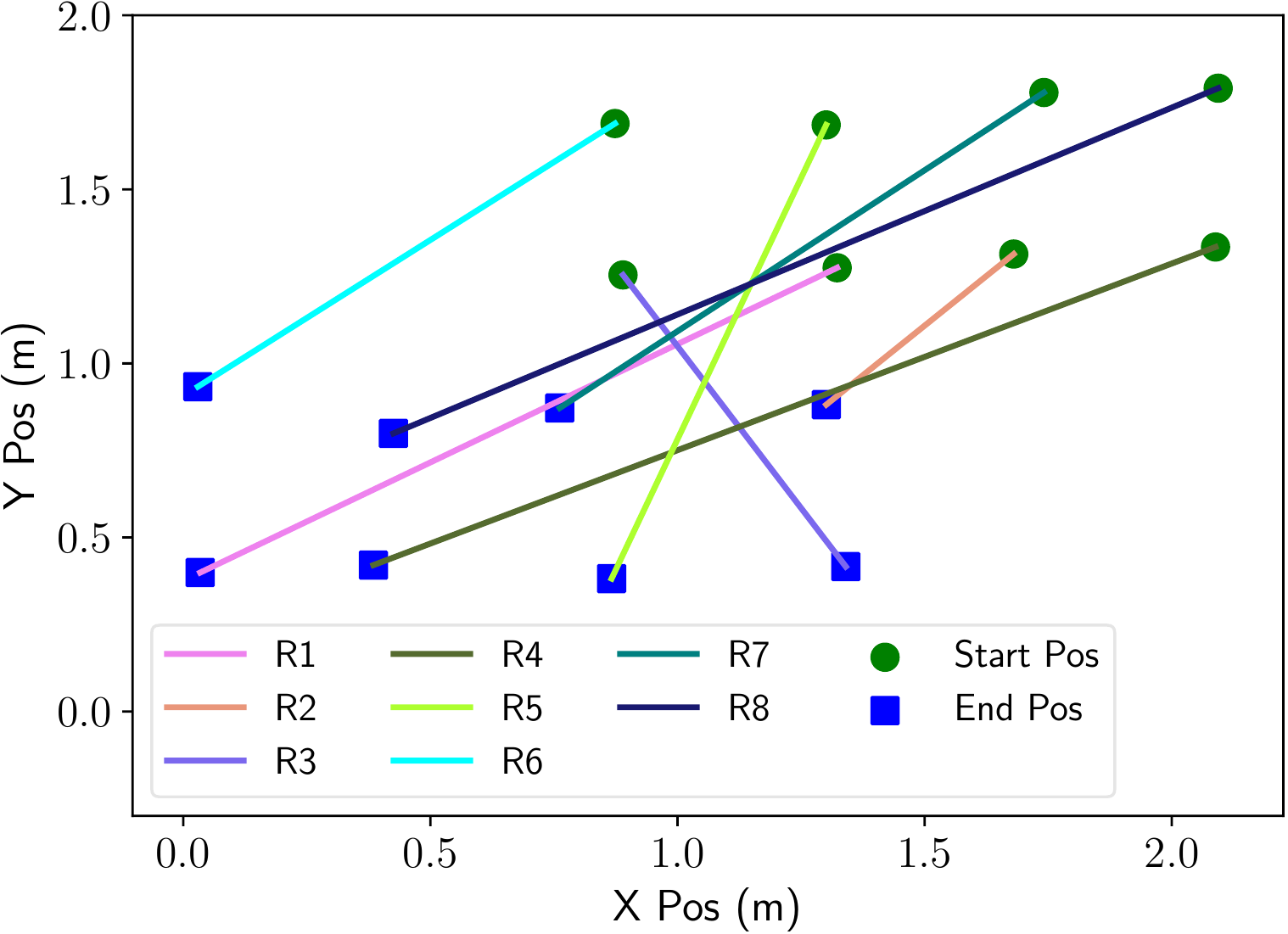}
		\caption{Random Assignment}
	\end{subfigure}
	\\
	\begin{subfigure}{.32\textwidth}
		\captionsetup{width=0.9\textwidth}
		\centering
        \includegraphics[width=\textwidth]{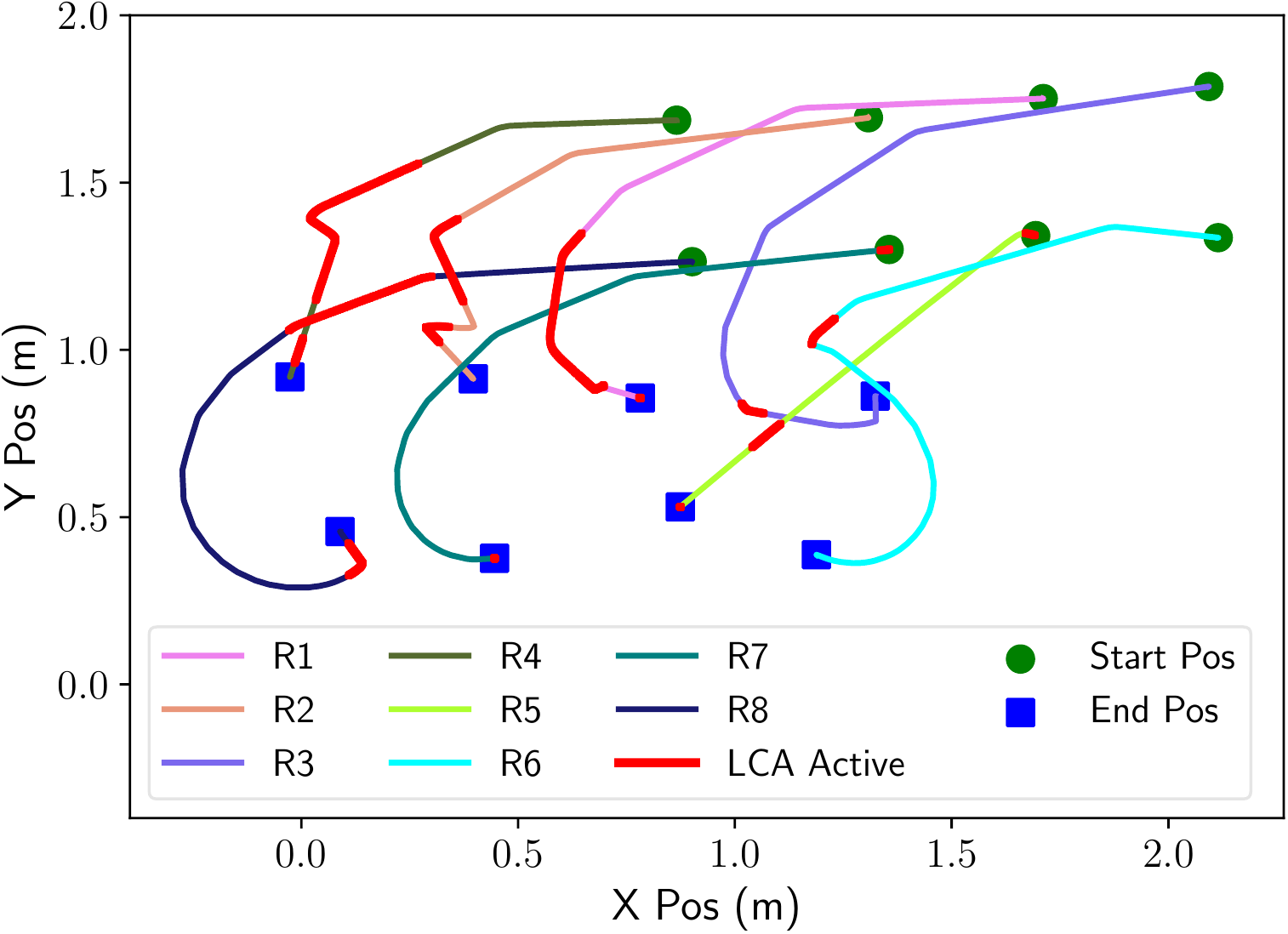}		
		\caption{Optimal}
		\label{fig:kh4_trajectory_sub_d}
	\end{subfigure}%
	\begin{subfigure}{.32\textwidth}
		\captionsetup{width=0.9\textwidth}
		\centering
        \includegraphics[width=\textwidth]{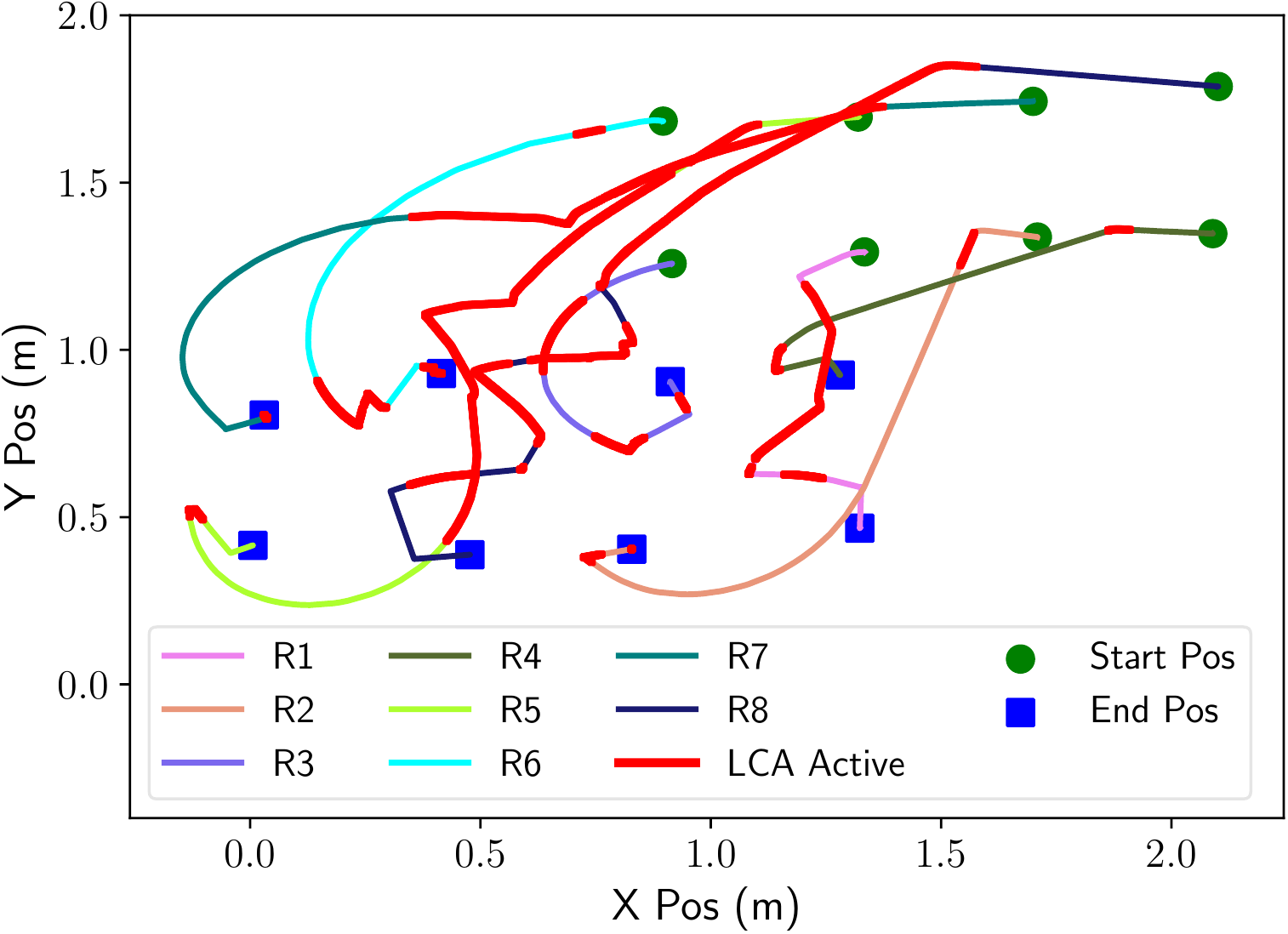}		
		\caption{CATA}
	\end{subfigure}
	\begin{subfigure}{.32\textwidth}
		\captionsetup{width=0.9\textwidth}
		\centering
        \includegraphics[width=\textwidth]{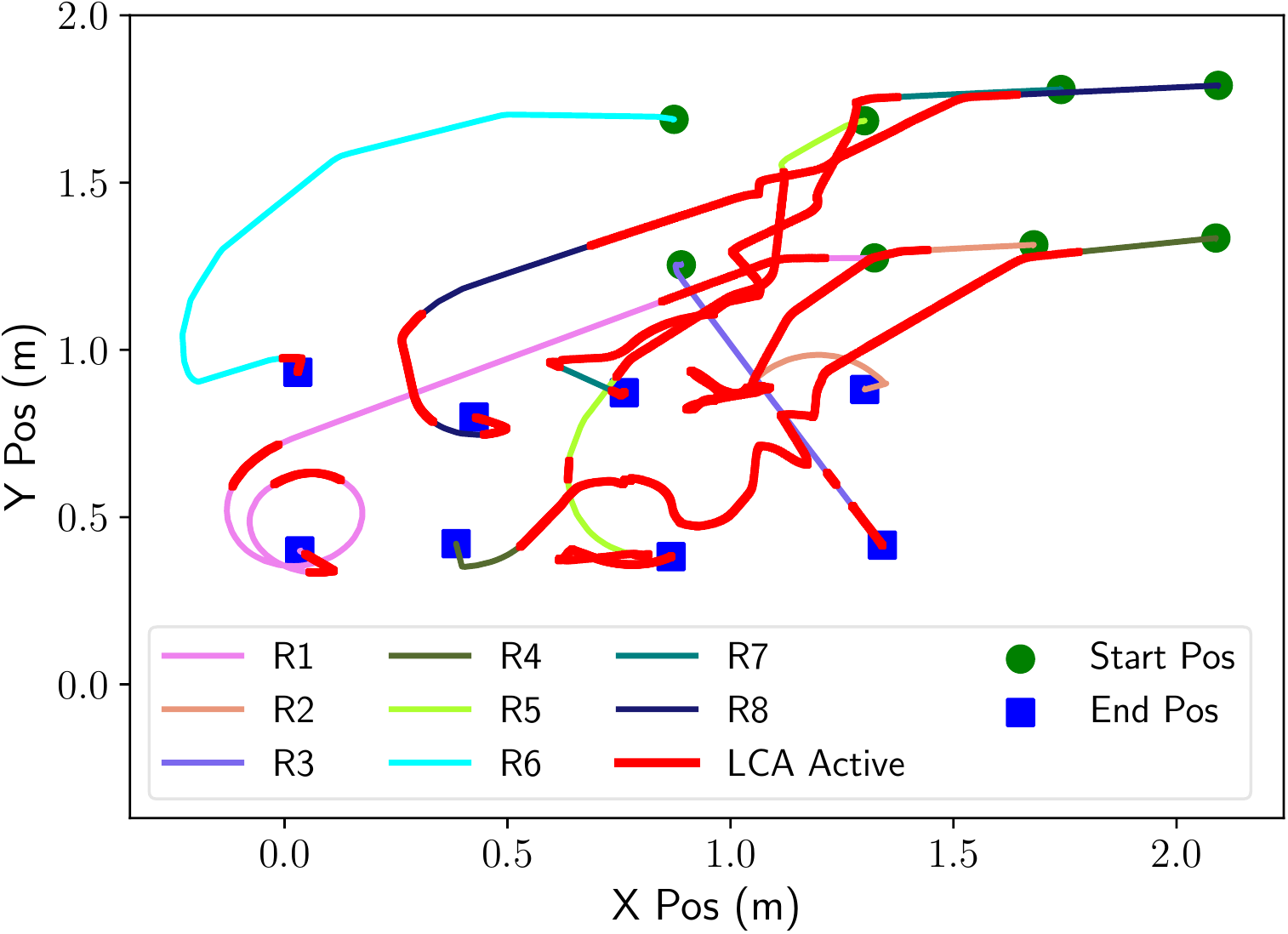}
		\caption{Random Assignment}
	\end{subfigure}
\caption{Top row indicating the trajectories considered during assignment and bottom row indicating trajectories taken by 8 KheperaIV robots under three experimental configurations}
\label{fig:kh4_trajectory}
\end{figure*}

We simulate 100 trials for each setup with $N=9$ and $N=25$. We monitor three
types of collision incidents throughout the mission: 1) any robot enters the
collision cone of any other robot 2) two robots enter each other's safety zone
without entering the collision cone 3) multiple robots enter each other's
safety zone without entering the collision cone. We refer to these scenarios
as \textit{avoidance}, \textit{maintain-one}, and \textit{maintain-multi},
respectively. The occurrence rate of these collision incidents is a good
indicator of the complexity level that local collision avoidance needs to
handle. Figure~\ref{fig_collision_incidents} shows the simulation results,
where CATA largely reduced all three types of collision incidents. It is worth
noting that as the number of robots increases, the occurrence of collision
incidents increased substantially when using CBAA. On the contrary, CATA
managed to limit this increase. In addition, the reduction of
\textit{maintain-multi} incidents is of particular interest because most of
the existing local collision avoidance methods perform poorly when handling
multi-robot interaction in real-life scenarios. We also observed deadlocks in
some trials, when collision avoidance simply failed. The dead lock occurrence
rate when $N=9$ and $N=25$ are summarized in Table~\ref{table_1}. As the total
number of robots increases, the grid setup generates more deadlocks than the line
setup because of the stopping robot scenarios explained in
Section~\ref{section_three_sub_c}. It might be tempting to always choose a line
setup over a grid setup, however, that quickly becomes unrealistic for practical
applications. Here we demonstrate CATA significantly reduced deadlocks for
both the grid setup and the line setup. Out of one hundred trials, fewer than 10 trials
failed using CATA, while 52 trials failed using CBAA for the grid setup with $N=25$.

Box plots of the mission completion steps of successful trials are presented
in Fig.~\ref{fig_steps}. The average completion steps remained approximately
the same for CBAA and CATA. Intuitively, this indicates that CATA successfully
reduced collision incidents without lengthening the overall mission.


\section{EXPERIMENTAL VALIDATION}
\label{section_five}



The performance of CATA was studied using a small team of 8
KheperaIV~\cite{Soares2016a} robots. Our experimental platform consists of an
IR camera based optical tracking systems (Optitrack), a central communication
hub emulating the inter-robot communications, and 8 KheperaIV robots. The
central communication hub obtained the position from the tracking system and
emulated situated communication~\cite{Stoy2001}, where receivers of a message
are aware of the senders' position in their own reference frames. During the
experiments, all the robots ran an instance of the Buzz Virtual Machine
(BVM)~\cite{Pinciroli2015} and executed identical scripts. The script includes
task assignment, velocity control and local collision avoidance
algorithms. Different task assignment schemes are used for comparison,
including CATA, manual optimal assignment, and random assignment. We
used a simple integrator controller for velocity control, which receives a
target position and applies a piecewise function to determine the left and
right wheel velocity of the differential drive robot.  When the robots move
too close to each other, a light-weight collision avoidance algorithm
(LCA)~\cite{Shahriari} exerts a virtual force on the robots and deflects it
away from other moving robots and obstacles.

With eight predefined task locations in Fig.~\ref{fig:kh4_trajectory}, we
report the task assignment results (top row) and robot trajectories (bottom
row) with CATA (middle), optimal assignment (left) and random assignment
(right). The optimal assignment was specified by a human operator and the
random assignment was obtained by randomly associating tasks to robots. It can
be observed that CATA is capable of taking into account the potential
collisions and provides reasonably detangled assignments.  The non-holonomic
nature of the robots resulted in spiral-shaped trajectories, and due to
imperfections in position estimation, some robots, such as R5 in
Fig.~\ref{fig:kh4_trajectory_sub_d}, turned on the spot and executed a
straight trajectory. In all three sub-figures of the second row, we have
marked the local collision avoidance activity with red lines, which means that
at least one of the robots was too close to its neighbor and triggered the
local collision avoidance. We repeated the CATA based task
assignment experiment three times and obtained roughly identical assignments,
except for small changes in the trajectories as a result of communication
delays, positioning errors, and asynchronous script execution.

\section{CONCLUSIONS}
\label{section_six}

We proposed a collision-aware task assignment strategy that considers potential collisions during the bidding process. By shaping the local rewards of tasks with collision cones and addressing the stopping robot problem with receding collision horizon, we successfully mitigated the inter-robot collisions during the task assignment stage. As a result, local collision avoidance method handles less and simpler collision incidents. We empirically evaluated the
approach with simulations and reported significantly improved results under various configurations. We also implemented the algorithm in Buzz on real robots and presented the trajectories with different task assignment schemes. As KheperaIV are differential wheeled robots, the actuation constraints introduced more complicated collision scenarios than the simulation results. For future work, we plan to extend our work by adapting the approach for nonholonomic robots and eventually heterogeneous robotic system. 



\section*{APPENDIX}

Here we show that CATA guarantee 50\% optimality when local reward function satisfies the \textit{diminishing marginal gain} condition. 

\textit{Proof}: Each round of auction produces one globally highest bid. For notational convenience, we use the same symbol for both the round identifier and the ID of the robot that wins the auction at the corresponding round. In other words, robot $r_i$ won the auction at round $i$ with bid $b_{gi,i}$ and robot $r_j$ won the auction at round $j$ with bid $b_{gj,j}$. We assume $i < j$ for the rest of this section.

Because only the globally highest bid wins the auction, the following condition holds:
\begin{equation}
	b_{gi, i} \geq b_{jA(r_i), i} \hspace{0.4cm} \forall i,j \in {1,... N_R}
\label{equ_inequ_1}
\end{equation}
where $A$ is an assignment set that can be searched with either the robot ID or the task identifier, $A(r_i)$ gives the task that robot $r_i$ has won at round \textit{i}.
Because each robot only submits its local highest bid and its local reward function satisfies the \textit{diminishing marginal gain} condition, which means the bid that any robot can submit for any task monotonically decreases as the auction proceeds, the following condition holds:
\begin{equation}
	b_{gi, i} \geq b_{il, i} \geq b_{il, j}	
	 \hspace{0.4cm} \forall i \in {1,... N_R} \hspace{0.2cm} \forall l \in 
	\label{equ_inequ_2}
\end{equation}

Consider swapping the tasks of robot $r_i$ and $r_j$, their combined bid changes from $b_{gi,i} + b_{gj,j}$ to $b_{iA(r_j),j} + b_{jA(r_i),i}$. Because the inequalities in (\ref{equ_inequ_1}) and (\ref{equ_inequ_2}) hold, the new combined bid is upper bounded as below.
\begin{equation}
	b_{iA(r_j),j} + b_{jA(r_i),i} \leq b_{gi,i} + b_{gi,i} = 2b_{gi,i}
\end{equation}
And the largest improvement of the combined bid of robot $r_i$ and $r_j$ is achieved when
\begin{equation}
	b_{iA(r_j),j} = b_{jA(r_i),i} = b_{gi,i}
\label{equ_swap}
\end{equation} 

Now consider CATA provides a solution so that the objective value is 
\begin{equation}
CATA = \sum_{i=1}^{N}b_{gi, i}
\end{equation}
where $N$ refers to the maximum number of tasks that can be assigned. Since each robot can only take one task, any possible variation of the assignment should be in the form of a sequence of task swapping. The largest possible improvement of the objective value can be achieved after a specific sequence of task swapping while every task swapping satisfies the condition (\ref{equ_swap}). Therefore the optimal objective value (OOV) should satisfy

$$
OOV = \sum_{i=1}^{N_{swapped}/2}b_{iA(r_j), j} + \sum_{j=1}^{N_{swapped}/2}b_{jA(r_i),i} + \sum_{i=1}^{N_{unswappped}}b_{gi,i}
$$
$$
= 2\sum_{i=1}^{N_{swapped}/2}b_{gi,i} + \sum_{i=1}^{N_{unswapped}}b_{gi,i} 
$$
$$
\leq 2\sum_{i=1}^{N}b_{gi,i} = 2CATA
$$
where $N_{swapped}$ refers to the number of tasks that need to be swapped to achieve the largest possible improvement, and $N_{unswapped}$ refers to the rest of the assignments.

Thus, $CATA \geq OOV/2$, the 50\% optimality is guaranteed.

\section*{ACKNOWLEDGMENT}

We would like to thank NSERC for supporting this work under the NSERC Strategic Partnership Grant (479149). Simulations were performed using the computing clusters managed by Calcul Qu\'ebec and Compute Canada.


\end{document}